\documentclass[]{gtech}
\usepackage{amssymb}
\usepackage{multirow}
\usepackage{bigdelim}
\usepackage{todonotes}
\usepackage{longtable}
\usepackage{tabularray}
\usepackage{wrapfig}
\usepackage[most]{tcolorbox} 
\usepackage{xcolor}
\usepackage{url}
\usepackage{enumitem}
\usepackage{hyperref}
\usepackage{float}

\renewcommand{\title}[1]{\newcommand{\titlelist}{{\huge\fontfamily{optimistic}\selectfont #1}}}

\usepackage{arydshln}

\definecolor{CQColor}{rgb}{0.0,0.0,1.0} % color for Aaron

\definecolor{TSColor}{rgb}{0.5,0.0,0.8} % color for Aaron

\definecolor{CQRColor}{rgb}{1.0,0.0,1.0} % color for Aaron

\usepackage{colortbl}
\usepackage{amssymb}
\usepackage{pifont}
\usepackage{booktabs,multirow}
\usepackage{makecell}
\usepackage{tabulary}
\usepackage{fontawesome5}
\usepackage{bbding}
\usepackage{multicol}

\newlength\savewidth

\title{\textcolor[HTML]{0369ff}{GO-MLVTON}: Garment Occlusion-Aware Multi-Layer Virtual Try-On with Diffusion Models}
\author[]{
Yang Yu$^{\dagger,\diamond}$, 
Yunze Deng$^{\dagger,\diamond}$, 
Yige Zhang$^\dagger$, 
Yanjie Xiao$^\dagger$, 
Youkun Ou$^\dagger$, 
Wenhao Hu$^\dagger$, 
Mingchao Li$^\dagger$, 
Bin Feng$^{\dagger,1}$, 
Wenyu Liu$^\dagger$, 
Dandan Zheng$^{\ddagger,2}$, 
Jingdong Chen$^{\ddagger,2}$
}

\contribution[]{$^\diamond$Contribute equally \\$^\dagger$School of EIC, Huazhong University of Science and Technology, Wuhan, China \\$^\ddagger$Inclusion AI, Ant Group, Beijing, China 
\\
$^1$The first corresponding author (fengbin@hust.edu.cn) \\
$^2$The second corresponding author (yuandan.zdd@antgroup.com, jingdongchen.cjd@antgroup.com)}

\abstract{\fontsize{11pt}{12pt} \textit{
Existing image-based virtual try-on (VTON) methods primarily focus on single-layer or multi-garment VTON, neglecting multi-layer VTON (ML-VTON), which involves dressing multiple layers of garments onto the human body with realistic deformation and layering to generate visually plausible outcomes. 
The main challenge lies in accurately modeling occlusion relationships between inner and outer garments to reduce interference from redundant inner garment features.
To address this, we propose GO-MLVTON, the first multi-layer VTON method, introducing the Garment Occlusion Learning module to learn occlusion relationships and the StableDiffusion-based Garment Morphing \& Fitting module to deform and fit garments onto the human body, producing high-quality multi-layer try-on results.
Additionally, we present the MLG dataset for this task and propose a new metric named Layered Appearance Coherence Difference (LACD) for evaluation.
Extensive experiments demonstrate the state-of-the-art performance of GO-MLVTON. Project page: \href{https://upyuyang.github.io/go-mlvton/}{https://upyuyang.github.io/go-mlvton/}.
}
}

\date{Feb 7, 2026\vspace{2mm}}

\begin{document}
\maketitle

\section{Introduction}
\label{sec:Introduction}

Image-based virtual try-on (VTON), as a prominent task in AIGC, aims to synthesize a realistic image of a person wearing the target garments while preserving identity features and garment details. Although significant advances have been made in VTON, existing methods focus solely on either single-garment try-on (SG-VTON)~\citep{han2019clothflow, kim2024stableviton, zeng2024cat, chong2024catvton, wang2025mv} or multi-garment try-on (MG-VTON)~\citep{zhu2024m, zhang2024mmtryon, choi2025controllable}, overlooking multi-layer garment try-on (ML-VTON), which is also essential in practical applications. ML-VTON aims to generate high-fidelity try-on results by layering multiple garments in a physically plausible manner, accurately simulating both the occlusion relationships between inner and outer layers and the realistic draping of garments on the human body.

A straightforward solution to ML-VTON is repeatedly applying SG-VTON method, but this requires manually providing try-on area masks for each garment layer, making it highly labor-intensive. More critically, it fails to model occlusion relationships among layers, where redundant features from occluded inner garments can adversely affect the network, causing details misalignment or unrealistic visual effects.

Therefore, this paper proposes \textbf{G}arment \textbf{O}cclusion-aware \textbf{M}ulti-\textbf{L}ayer \textbf{V}irtual \textbf{T}ry-\textbf{On} (\textbf{GO-MLVTON}). To address the challenge of learning relationships among garment layers, we meticulously design the Garment Occlusion Learning (GOL) module to learn the occlusion relationships of different layers to refine the inner garment features. Then we utilize the StableDiffusion-based Garment Morphing \& Fitting (GMF) module to deform and fit layered garments onto the human body. In this way, GO-MLVTON achieves high-fidelity try-on results with realistic layering and occlusion.
%, the first 2-layer-garment virtual try-on method for the ML-VTON task.

\begin{figure}[t]
    \centering
    \captionsetup{justification=centering}
    \includegraphics[width=1.0\linewidth]{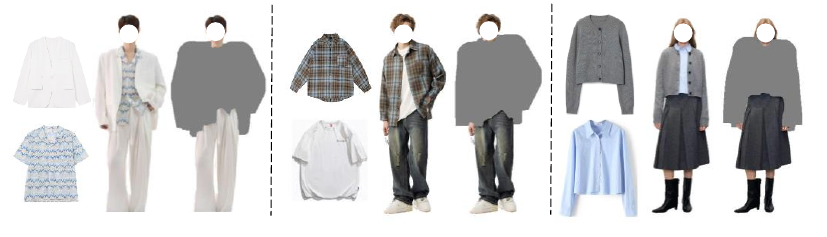}
    \caption{Examples from the MLG dataset.
    }
    \label{fig: dataset}
\end{figure}

Besides that, the lack of a dedicated dataset and evaluation metrics for the ML-VTON task is another major challenge. Therefore, we introduce the first multi-layer garment dataset, named MLG, consisting of 3,538 samples. As shown in Fig.\ref{fig: dataset}, each sample contains a quadruplet of: (1) an inner garment; (2) an outer garment; (3) a human image wearing both garments; and (4) a clothing-agnostic image of the same person with the upper-body region masked. We also propose the Layered Appearance Coherence Difference (LACD) for evaluation. 
For more details, please refer to Sec.\ref{sec: MLGdataset and LACD}.
%To evaluate the visual coherence and layering quality of generated results, w
% Existing image-based virtual try-on public datasets only support SG-VTON or MG-VTON, lacking both multi-layer garment instances and corresponding evaluation metrics.
% This metric computes pixel-level L2 differences between the ground truth and the generated image across different garment layers, while assigning higher weight coefficients to the connecting regions between inner and outer layers to emphasize structural consistency and detail fidelity in multi-layer garment generation.
We conduct extensive experiments using GO-MLVTON on MLG. Both quantitative and qualitative results demonstrate that our method outperforms existing VTON methods in the ML-VTON task, producing higher-quality try-on results.

In summary, our contributions are: (1) We introduce GO-MLVTON, the first end-to-end model designed for the ML-VTON task. The GOL module and the GMF module are meticulously designed to refine inner garment features by leveraging occlusion relationships and to generate realistic try-on results, respectively. (2) We collect a multi-layer garment dataset named MLG, and propose a novel evaluation metric named LACD, filling a gap in the ML-VTON field. (3) Extensive experiments conducted on MLG demonstrate that GO-MLVTON exhibits superiority in the ML-VTON task.

\section{Related Work}
\label{sec:RelatedWork}

\textbf{Single-garment Virtual Try-on.}
Early methods~\citep{han2019clothflow, choi2021viton} adopt a two-stage GAN-based paradigm, first warping the garment to align with the target pose, then synthesizing the final result via a generator. However, these methods often yield unsatisfactory results when dealing with complex poses, limiting real-world applicability. Recently, diffusion models, leveraging their generative capabilities, have been introduced to VTON with impressive performance. 
For instance, StableVITON~\citep{kim2024stableviton} and CAT-DM~\citep{zeng2024cat} both utilize a ControlNet~\citep{zhang2023adding} architecture to extract multi-scale garment features and align them with human body features through cross-attention blocks, generating high-quality try-on results. 
%SG-VTON is designed for VTON of a single reference garment. 
%Given a target person image and a reference garment image, SG-VTON generates realistic images of the target wearing the specified clothing.
% Focusing on a different challenge, TPD[25] preserves as much identity information as possible by predicting an accurate inpainting area for each person-garment image pair. 
CATVTON~\citep{chong2024catvton} achieves a lightweight solution by spatially concatenating the garment and person image. 
%Furthermore, MV-VTON~\citep{wang2025mv} introduces a view-adaptive selection mechanism to dynamically fuse front and back view garment features based on human poses, generating high-fidelity multi-view try-on results.

\textbf{Multi-garment Virtual Try-on.} Several studies have focused on MG-VTON. M\&M VTO~\citep{zhu2024m} achieves realistic results through a single-stage diffusion model with text-guided layout control. Another framework, MMTryon~\citep{zhang2024mmtryon}, utilizes Multi-Modal Instruction Attention and Multi-Reference Texture Attention mechanism, enabling detailed text-instruction control alongside multi-garment support. To handle an arbitrary number of garments, AnyDressing~\citep{li2025anydressing} introduces a shared Garment-Specific Feature Extractor and integrates the garment features into the denoising process via a Dressing-Attention module. Although multiple garments are supported, they do not involve complex overlaps or occlusions, which constrains the applicability of these methods in ML-VTON scenarios, constituting the central problem that our work seeks to address.
%BootComp~\citep{choi2025controllable} employs two separate diffusion models to extract features from garments and human bodies, respectively, fusing them through the self-attention mechanism. 

\begin{figure*}[t]
    \centering
    \captionsetup{justification=centering}
    \includegraphics[width=1.0\linewidth]{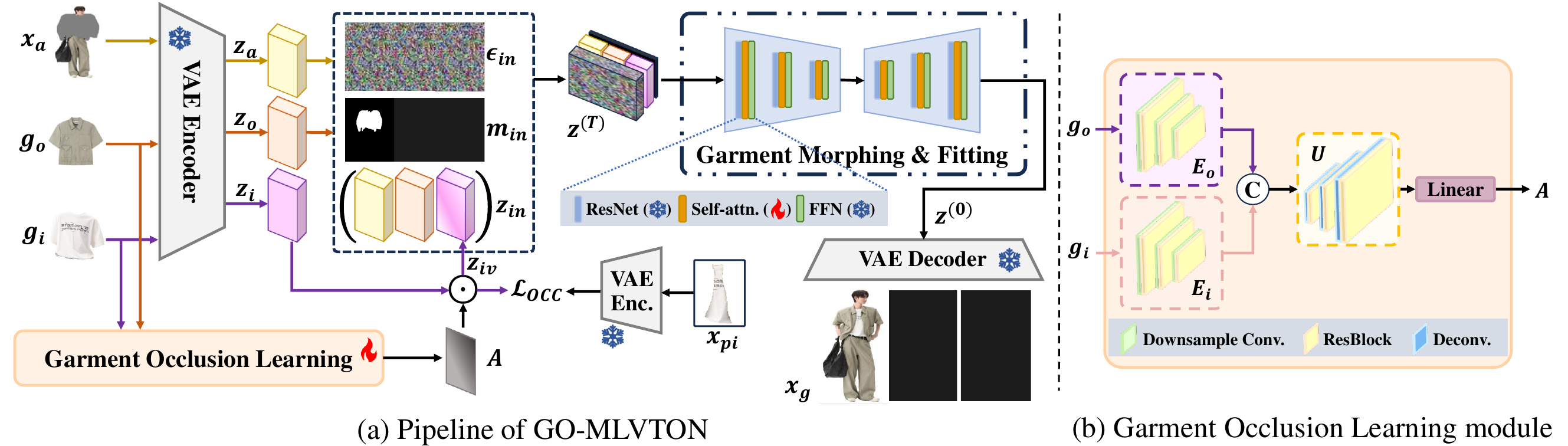}
    \caption{(a) Pipeline of GO-MLVTON. (b) Details of the Garment Occlusion Learning module.
    }
    % \vspace{-5mm}
    \label{fig: Pipeline}
\end{figure*}

\section{MLG Dataset And Layered Appearance Coherence Difference}
\label{sec: MLGdataset and LACD}

Public datasets for image-based VTON, such as VITON-HD\citep{choi2021viton} and DressCode\citep{morelli2022dress}, are limited to single-layer garment VTON. 
To overcome this, we propose MLG, the first dataset for ML-VTON. 
We first collect appropriate human images and corresponding inner and outer garment images from Taobao\footnote{https://taobao.com} e-commerce platforms. 
Since some samples lack standalone images of the inner or outer garments, we then use model-wearing images as sources and apply the Segment Anything Model (SAM)\citep{kirillov2023segment} to segment and extract inner and outer garments separately. Finally, we utilize the SCHP\citep{li2020self} model to obtain human parsing maps for constructing clothing-agnostic person images.

%MLG contains 3,538 samples, with each sample consisting of three components: 1) an outer garment, 2) an inner garment and 3) a human image wearing the corresponding inner and outer garments. The examples of the MLG dataset are illustrated in Fig.\ref{fig: dataset}. 

Furthermore, to address the limitation of existing metrics focusing solely on overall image quality without fine-grained assessment of specific regions in generated results, we propose the metric LACD. LACD computes weighted pixel-level $L2$ differences between ground truth and generated images across garment layers, especially emphasizing transitional regions between adjacent layers. For $N$ garment layers in a generated image, let $A_i$, $B_i \in A_i$, and $C_i = A_i \setminus B_i$ denote the $i$-th layer region, the connecting region with the $(i+1)$-th layer, and the interior region, respectively.
The layered consistency difference for the $i$-th layer is defined as:

\begin{equation}
    \begin{aligned}
         lacd_i=\lambda_1\sum_{j\in B_i}\left \| x_{gt}^{(i, j)} - x_{gen}^{(i, j)}\right \|  _2 + \sum_{k\in C_i}\left \| x_{gt}^{(i, k)} - x_{gen}^{(i, k)}\right \|  _2, 
    \end{aligned}
\end{equation}
where $x_{gt}^{(i,j)}$ and $x_{gen}^{(i,j)}$ denote the color values at the $j$-th pixel position in the $i$-th layer of the ground truth and generated images, respectively. $\lambda_1 = 3$ is a weighting coefficient that emphasizes the importance of coherence in connecting regions.
The overall LACD is computed as the average of all layer differences: $LACD = \frac{1}{N}\sum_{i=1}^{N}{lacd_i}$.
By explicitly reinforcing coherence in inter-layer connecting regions through fixed weighting, this metric effectively measures the structural preservation and visual consistency of the garment areas in the generated images, giving higher priority to the naturalness of layer transitions and occlusions.

% \begin{equation}
%     LACD = \frac{1}{N}\sum_{l=1}^{N}{lacd_i}, 
% \end{equation}

\section{Method}
\label{sec: Method}

ML-VTON aims to generate a realistic try-on image $x_g$ from a person image $x_p$, inner garment $g_i$, and outer garment $g_o$, ensuring natural layering and plausible occlusion. To address this, we propose \textbf{GO-MLVTON}, an end-to-end method with two key components: the GOL module for learning garment occlusion relationships and the GMF module for modeling garment deformation and human-garment interactions to produce high-quality results.

Specifically, we regard this task as an exemplar-based image inpainting problem, where garment $g_i$ and $g_o$ serve as references to fill the specified masked region in the clothing-agnostic person image $x_a$ (obtained by the element-wise multiplication of $x_p$ and the upper-body clothing mask $M$), synthesizing a realistic overlapped dressing effect. Fig.\ref{fig: Pipeline}(a) shows that $x_a $, $g_o$, $g_i\in \mathbb{R}^{3 \times H \times W}$ are first processed by a VAE Encoder $\mathrm{\varepsilon}$ to extract latent features:
\begin{equation}
    z_a = \mathrm{\varepsilon}(x_a), z_o = \mathrm{\varepsilon}(g_o), z_i = \mathrm{\varepsilon}(g_i), 
\end{equation}
where $z_a$, $z_o$, $z_i\in \mathbb{R}^{4 \times \frac{H}{8} \times \frac{W}{8}}$. 
% The GOL module then models the spatial occlusion relationship between $g_i$ and $g_o$, producing the attention map $A$, which is used to refine the inner garment feature $z_i$ by enhancing visible regions and suppressing occluded areas:
The GOL module then models the spatial occlusion relationships between $g_i$ and $g_o$, generating an attention map $A = GOL(g_i, g_o)\in \mathbb{R}^{1 \times \frac{H}{8} \times \frac{W}{8}}$ to refine $z_i$ by enhancing visible regions and suppressing occluded areas. 
% Subsequently, the refined inner garment feature $z_{iv}$ is obtained by element-wise multiplication of the attention map $A$ and the inner latent feature $z_i$:
Subsequently, the refined inner garment feature $z_{iv}$ is obtained by weighting $z_i$ with the attention map $A$: $z_{iv} = A \odot z_i$. Here, $\odot$ represents the element-wise multiplication. Next, the following features are concatenated spatially: $z_{in} = [z_a, z_o, z_{iv}]$, $m_{in} = [m_a, m_0, m_{0}]$, and $\epsilon_{in} = [\epsilon_a, \epsilon_o, \epsilon_{iv}]$, 
% \begin{equation}
%     \begin{aligned}
%     z_{in} &= [z_a, z_o, z_{iv}], m_{in} = [m_a, m_0, m_{0}], 
%     \\
%     \epsilon_{in} &= [\epsilon_a, \epsilon_o, \epsilon_{iv}],
%     \end{aligned}
% \end{equation}
where $[\cdot,\cdot, \cdot]$ is spatial concatenation; $\mathrm{\epsilon}_a$, $\mathrm{\epsilon}_o$, $\mathrm{\epsilon}_{iv}$ are the random Gaussian noise; $m_a$ is the binary mask indicating the inpainting region; $m_0$ is the all-zero mask. 
$z_{in}$, $\mathrm{\epsilon}_{in}$, and $m_{in}$ are then concatenated as input to the GMF module, which models garment-body interactions and processes garment deformation to produce final result $z^0$. Finally, $z^0$ is decoded by VAE Decoder $D$: $x_g = D(GMF(z^T))$, where $z^T=z_{in} \copyright \epsilon_{in} \copyright m_{in}$, and $\copyright$ is the channel concatenation.
% \begin{equation}
% \left\{\begin{matrix}
% z^T=z_{in} \copyright \epsilon_{in} \copyright m_{in} \\
% x_g = D(GMF(z^T))
% \end{matrix}\right. .
% \end{equation}

\subsection{Garment Occlusion Learning Module}

This module explicitly models the spatial occlusion between inner and outer garments, refining $z_i$ by suppressing occluded regions and preserving visible information, as shown in Fig.\ref{fig: Pipeline}(b). To capture global garment semantics, two identical encoders ($E_o$ and $E_i$) are employed to process $g_o$ and $g_i$, obtaining garment representations $f_o$ and $f_i$. Each encoder has $M$ ($M$ = 5) layers, comprising a downsampling convolution and two residual blocks.

Subsequently, $f_o$ and $f_i$ are concatenated along the channel dimension and processed by a mapping network $U$ and a linear layer to learn their occlusion relationships, producing a spatial attention map $A$. Furthermore, we introduce the supervised loss function $\mathcal{L}_{OCC}$ to ensure the effectiveness, and the overall process of GOL can be written as:  
\begin{equation}
\left\{\begin{matrix}
A=Linear(U(E_o(g_o) \copyright E_i(g_i))) \\
\mathcal{L}_{OCC} =  \| \mathrm{\varepsilon}(x_{pi}) - z_{iv} \|  _2
\end{matrix}\right. ,
\end{equation}
where $x_{pi} \in \mathbb{R}^{3 \times H \times W}$ denotes the inner garment regions cropped from the ground truth image.

\subsection{Garment Morphing \& Fitting Module}

This module leverages a StableDiffusion-based UNet to synthesize the final output from the concatenated input latent features. 
Garment geometric morphing is progressively achieved through the UNet's hierarchical modeling, where self-attention blocks at each scale can model semantic relationships between the human body and layered garments. These blocks align garment features with pose and shape cues from human latent encoding, while human features constrain garment deformation for a realistic fit. This ensures seamless fusion of the body and multi-layer attire, ultimately producing high-fidelity try-on results. During this process, the denoising loss is calculated as:
\begin{equation}
\mathcal{L}_{GMF} = \mathbb{E}_{\epsilon \sim \mathcal{N}(0,1),t}[\left \| \epsilon - \epsilon_{\theta } (z^t,t) \right \|_2^2 ],
\end{equation}
where $z^t$ is the input of GMF at the $t$-th denosing step, $t\in{\{1, ..., T\}}$ and $\epsilon \sim \mathcal{N}(0,1)$ is the random Gaussian noise. Thus, the overall loss function can be defined as $\mathcal{L} = \mathcal{L}_{GMF} + \lambda_2\mathcal{L}_{OCC}$, where $\lambda_2 = 0.1$.

\section{Experiments}
\label{sec: Experiments}

\subsection{Datasets and Implementation Details}

\begin{table}[t]\centering
\def\arraystretch{1.2}
\captionsetup{justification=centering}
\small
\tabcolsep 1.5pt
    \begin{tabular}{lc cc cccccc}
        \toprule
        \multicolumn{2}{c}{Method} && Year && FID$\downarrow$ & KID$\downarrow$ & SSIM$\uparrow$ & LPIPS$\downarrow$ & LACD$\downarrow$ \\
        \cmidrule{1-2} \cmidrule{4-4} \cmidrule{6-10}
        \multicolumn{2}{c}{CAT-DM~\citep{zeng2024cat}} &&  CVPR24 && 32.36 & 5.97 & \underline{0.845} & 0.138 & 0.719 \\
        \multicolumn{2}{c}{MV-VTON~\citep{wang2025mv}} &&  AAAI25 && 36.31 & 7.60 & 0.830 & 0.175 & 0.973 \\
        \multicolumn{2}{c}{CATVTON~\citep{chong2024catvton}} && ICLR25 && \underline{30.54} & \underline{4.61} & 0.841 & \underline{0.127} & \underline{0.626} \\
        \cmidrule{1-2} \cmidrule{4-4} \cmidrule{6-10}
        \multicolumn{2}{c}{Ours} && - && \textbf{22.82} & \textbf{0.35} & \textbf{0.858} & \textbf{0.108} & \textbf{0.623}\\ 
        \bottomrule
    \end{tabular}
\vspace{-2mm}
\captionof{table}{Quantitative comparisons on the MLG dataset.}
\vspace{-2mm}
\label{tab:mlg}
\end{table}

We conduct experiments on our MLG dataset, which is randomly split into 2,783 training pairs and 755 testing pairs. 
The GMF module is based on StableDiffusion v1.5~\citep{rombach2022high}, initialized with pretrained weights from InstrucPix2Pix~\citep{brooks2023instructpix2pix}, and cross-attention blocks are removed following CATVTON~\citep{chong2024catvton}. 
The model is optimized using AdamW~\citep{loshchilov2017decoupled} with learning rate $1e-5$ and batch size 8, training only the GOL module and GMF's self-attention blocks.
During training, 10\% conditional dropout is adopted, and inference uses classifier-free guidance~\citep{ho2022classifier}, where unconditional noise prediction is achieved by setting $z_o$ and $z_{iv}$ to zero, with a guidance scale of $s=2.5$.

\begin{figure}[t]
    \centering
    \captionsetup{justification=centering}
    \includegraphics[width=0.7\linewidth]{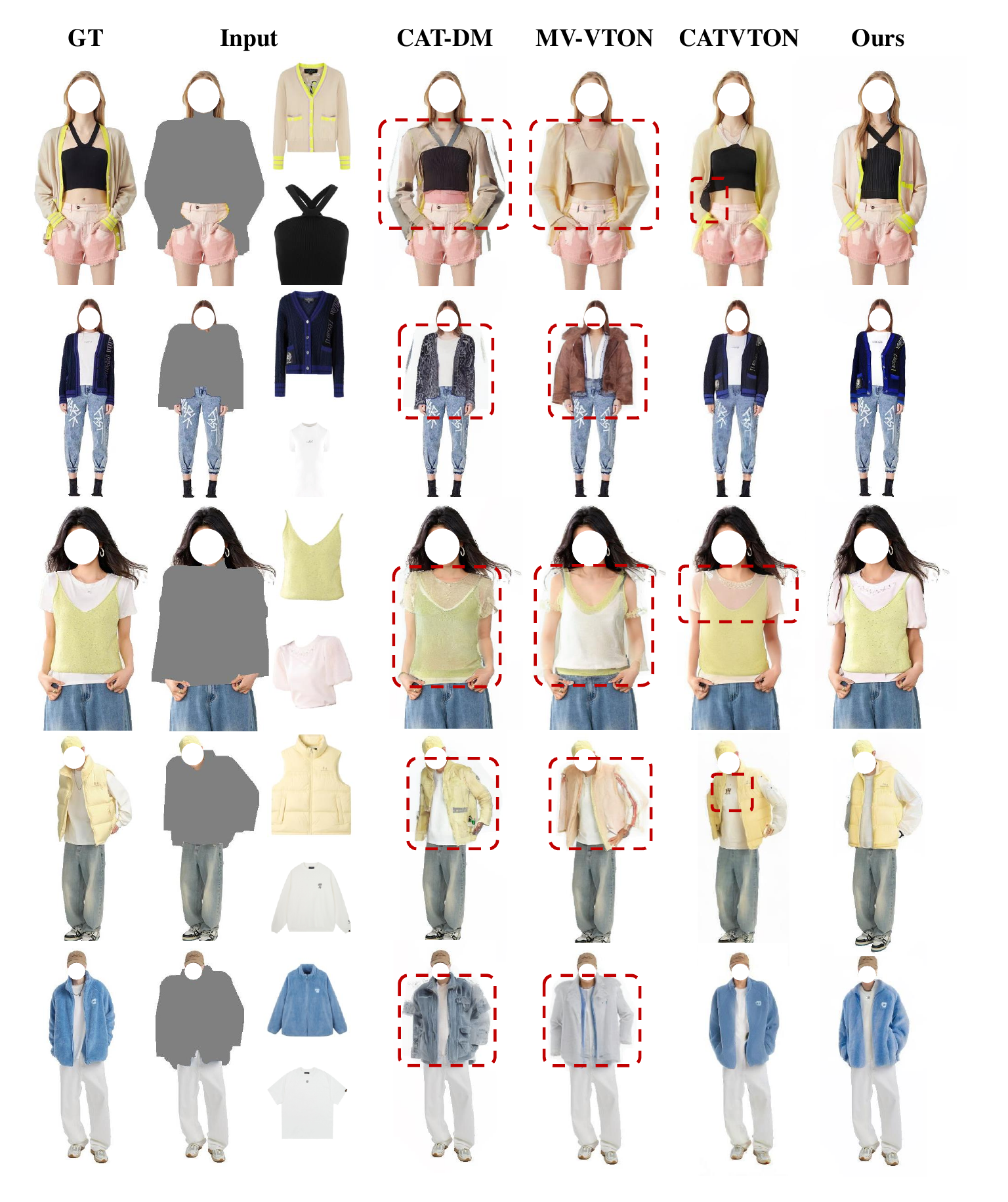}
    \caption{Qualitative Results. Red boxes indicate areas that exhibit visually unrealistic effects or details misalignment.
    }
    % \vspace{-5mm}
    \label{fig: results}
\end{figure}

\subsection{Evaluation Metrics and Comparing Settings}
Five evaluation metrics are employed to quantitatively assess our method: Structural Similarity (SSIM)~\citep{wang2004image}, Learned Perceptual Image Patch Similarity (LPIPS) ~\citep{johnson2016perceptual}, Frechet Inception Distance (FID)~\citep{heusel2017gans}, Kernel Inception Distance (KID)~\citep{binkowski2018demystifying}, and the proposed LACD. Besides, we compare our method with CAT-DM~\citep{zeng2024cat}, MV-VTON~\citep{wang2025mv} and CATVTON~\citep{chong2024catvton}. Since these methods are only applicable to single-layer VTON task, we divide the process into two steps: at each step, the mask and the clothing-agnostic person image, both corresponding to the target garment layer, serve as inputs to progressively produce the inner layer outcomes followed by the overall layered results.

\subsection{Qualitative Results}

\begin{table}[t]\centering
\def\arraystretch{1.2}
\small
\captionsetup{justification=centering}
\tabcolsep 1.5pt
    \begin{tabular}{lc cccccc}
        \toprule
        \multicolumn{2}{c}{Model Settings} && FID$\downarrow$ & KID$\downarrow$ & SSIM$\uparrow$ & LPIPS$\downarrow$ & LACD$\downarrow$ \\
        \cmidrule{1-2} \cmidrule{4-8}
        \multicolumn{2}{c}{Base} && \underline{25.71} & 1.23 & 0.828 & \underline{0.120} & \underline{0.625} \\
        \multicolumn{2}{c}{Base + GOL} && 25.81 & \underline{1.00} & \underline{0.847} & 0.124 & 0.868 \\
        \multicolumn{2}{c}{Base + GOL + $\mathcal{L}_{OCC}$} && \textbf{22.82} & \textbf{0.35} & \textbf{0.858} & \textbf{0.108} & \textbf{0.623} \\
        \bottomrule
    \end{tabular}
\vspace{-2mm}
\captionof{table}{Ablation Studies of the GOL module.}
\vspace{-2mm}
\label{tab: GOL_ablation}
\end{table}

\begin{table}[t]\centering
\def\arraystretch{1.2}
\small
\captionsetup{justification=centering}
\tabcolsep 1.5pt
    \begin{tabular}{lc cccccc}
        \toprule
        \multicolumn{2}{c}{Guidance Scale} && FID$\downarrow$ & KID$\downarrow$ & SSIM$\uparrow$ & LPIPS$\downarrow$ & LACD$\downarrow$ \\
        \cmidrule{1-2} \cmidrule{4-8}
        \multicolumn{2}{c}{0.0} && 39.89 & 7.93 & 0.809 & 0.177 & 1.256 \\
        \multicolumn{2}{c}{2.5} && \textbf{22.82} & \textbf{0.35} & \textbf{0.858} & \textbf{0.108} & \textbf{0.623} \\
        \multicolumn{2}{c}{5.0} && \underline{27.18} & \underline{2.84} & \underline{0.850} & \underline{0.119} & \underline{0.693} \\
        \bottomrule
    \end{tabular}
\vspace{-2mm}
\captionof{table}{Ablation Studies of the Classfier-Free Guidance.}
\vspace{-2mm}
\label{tab:CFG_ablation}
\end{table}

Fig.\ref{fig: results} presents the qualitative comparisons between GO-MLVTON and several state-of-the-art single-layer try-on methods on MLG. It can be observed that the proposed method effectively models reasonable occlusion relationships between the inner and outer garments, producing multi-layer try-on results with clear layering and a natural appearance. In contrast, single-layer methods struggle with multi-layer scenarios, often leading to details misalignment (\textit{e.g.}, CATVTON in the fourth row of Fig.\ref{fig: results}) and unrealistic visual effects due to interference from occluded inner garment features.

\subsection{Quantitative Results}

Tab.\ref{tab:mlg} shows that our method surpasses CAT-DM, MV-VTON, and CAT-VTON in all metrics. This consistent improvement can be attributed to our method's capability to effectively refine the features of the inner garments and suppress interference from redundant information in occluded regions.

\subsection{Ablation Studies}

\textbf{Garment Occlusion Learning Module.} 
To validate the effectiveness of the GOL module, we set our baseline model (Base) as directly using the unrefined inner garment feature $z_i$ without GOL and the supervision $\mathcal{L}_{OCC}$.
On this basis, we incrementally incorporate GOL (Base+GOL) and $\mathcal{L}_{OCC}$ (Base+GOL+$\mathcal{L}_{OCC}$). 
The results in Tab.\ref{tab: GOL_ablation} show that only GOL may retain occluded inner garment features due to the lack of supervision, affecting generation quality, while applying both GOL and $\mathcal{L}_{OCC}$ can effectively refine $z_{iv}$ by suppressing occluded features, enhancing VTON results.

\textbf{Classifier-Free Guidance.} To assess Classifier-Free Guidance (CFG) impact, we conduct inference experiments with guidance scales $s$ of 0.0, 2.5, and 5.0. Results in Tab.\ref{tab:CFG_ablation} show that both overly low and high scales degrade output quality, thus we set $s=2.5$ during inference.

\section{Conclusion}
\label{sec: Conclusion}
In this paper, we propose GO-MLVTON, the first ML-VTON method. We design the GOL module to refine inner garment features by suppressing occlusion interference while preserving visible area information. 
The GMF module processes these features with outer garment and human body features to achieve garment warping and interaction modeling, producing high-quality results.
Additionally, we collect a multi-layer garment dataset and propose the LACD metric to address data and evaluation gaps, with experiments showing GO-MLVTON's superiority in the ML-VTON task.

\section{Acknowledgement}
This work was supported by National Natural Science Foundation of China (No. 62376102) and Ant Group.

\label{sec: Conclusion}

% \footnotesize
\bibliographystyle{assets/plainnat}
\bibliography{paper}

\end{document}